\documentclass[conference]{IEEEtran}
\IEEEoverridecommandlockouts
\usepackage{cite}
\usepackage{amsmath,amssymb,amsfonts}
\usepackage{algorithmic}
\usepackage{graphicx}
\usepackage{textcomp}
\usepackage[table,xcdraw]{xcolor}
\usepackage{multirow}
\usepackage{authblk}
\usepackage{hyperref}
\def\BibTeX{{\rm B\kern-.05em{\sc i\kern-.025em b}\kern-.08em
    T\kern-.1667em\lower.7ex\hbox{E}\kern-.125emX}}
\begin{document}

\title{Multi-Modality Representation Learning for Antibody-Antigen Interactions Prediction}

\author{
    Peijin Guo\textsuperscript{\rm 1 $\dagger$},
    Minghui Li\textsuperscript{\rm 2 $\ddagger$},
    Hewen Pan\textsuperscript{\rm 1 $\dagger$},
    Ruixiang Huang\textsuperscript{\rm 2 $\ddagger$},
    Lulu Xue\textsuperscript{\rm 1 $\dagger$},\\
    Shengqing Hu\textsuperscript{\rm 3 $*$},
    Zikang Guo\textsuperscript{\rm 2 $\ddagger$},
    Wei Wan\textsuperscript{\rm 1 $\dagger$},
    Shengshan Hu\textsuperscript{\rm 1 $\dagger$}\\
    \footnotesize \texttt{\{gpj,minghuili,hewenpan,ruixiangh,lluxue,zikangguo,wanwei\_0303,hushengshan\}@hust.edu.cn}\\
    \texttt{hsqha@126.com}

     \thanks{2025 IEEE International Conference on Multimedia and Expo (ICME 2025), June 30 - July 4, 2025, Nantes, France.}
    
} 
 
\affil[1]{School of Cyber Science and Engineering, Huazhong University of Science and Technology, Wuhan, China}
\affil[2]{School of Software Engineering, Huazhong University of Science and Technology, Wuhan, China}
\affil[3]{Union Hospital, Tongji Medical College, Huazhong University of Science and Technology, Wuhan, China}

\maketitle

\begin{abstract}
   While deep learning models play a crucial role in predicting antibody-antigen interactions (AAI), the scarcity of publicly available sequence-structure pairings constrains their generalization. Current AAI methods often focus on residue-level static details, overlooking fine-grained structural representations of antibodies and their inter-antibody similarities. To tackle this challenge, we introduce a multi-modality representation approach that integates 3D structural and 1D sequence data to unravel intricate intra-antibody hierarchical relationships. By harnessing these representations, we present MuLAAIP, an AAI prediction framework that utilizes graph attention networks to illuminate graph-level structural features and normalized adaptive graph convolution networks to capture inter-antibody sequence associations. Furthermore, we have curated an AAI benchmark dataset comprising both structural and sequence information along with interaction labels. Through extensive experiments on this benchmark, our results demonstrate that MuLAAIP outperforms current state-of-the-art methods in terms of predictive performance. The implementation code and dataset are publicly available 
   at \url{https://github.com/trashTian/MuLAAIP} for reproducibility.
\end{abstract}


\section{Introduction}
Antibodies, as highly specific immunoglobulins, play a crucial role in recognizing and binding to specific antigens, initiating immune responses that either neutralize pathogens or mark them for clearance. Identifying antibodies with desired properties is a resource-intensive and time-consuming task, despite advancements in various experimental assays \cite{maynard2000antibody}. The vastness of the chemical space and the requirement for large quantities of purified antibodies further complicate this task. To address these challenges, deep learning-based approaches for predicting Antibody-Antigen Interactions (AAI) \cite{zhang2022predicting} have emerged as promising alternatives. The primary goal of these methods is to enhance the antibody screening process by predicting interaction strength (e.g., binding affinity) and specificity (e.g., neutralization) utilizing antibody structure or sequence data.

These methods typically fall into three categories: sequence-based, structure-based and multi-modality-based approaches. \textit{Sequence-based} methods either (1) conduct end-to-end representation learning and AAI prediction \cite{zhang2022predicting,li2024mvsf,wang2024abimmpred,yuan2023dg,wan2023pesi,yang2023area,huang2022abagintpre}, or (2) utilize protein language models \cite{9477085,lin2023evolutionary} or antibody language models \cite{luo2023bert2dab,olsen2022ablang,jing2024accurate} to derive antibody representations for downstream tasks like affinity prediction \cite{li2024mvsf,yuan2023dg}. While sequence-based methods prove valuable in the absence of structural data, the reliance solely on sequences poses inherent limitations due to the dominant role of three-dimensional structures in protein interactions \cite{dobson2003protein}. \textit{Structure-based} methods generally use graph neural networks (GNNs) to learn the residue-level structural representations for predicting AAI \cite{chinery2023paragraph,pittala2020learning}. However, these approaches frequently focus on individual residues, while disregarding fine-grained essential information, such as side-chain details, crucial for grasping antibody functionality.

Exploring the co-modeling of structure and sequence could unveil deeper functional insights. Previous work, leveraging a \textit{multi-modality} approach \cite{lu2021leveraging}, concatenated sequence and structural representations as node embeddings within a Graph Convolutional Network (GCN). However, this study overlooked sequence-structure dependencies, potentially missing functional similarities among proteins (\textit{e.g., }antibodies) that exhibit shared structural or sequential attributes \cite{zhang2022predicting,zheng2024progressive}.

In addition, the AAI prediction task faces two main bottlenecks related to real-world data: \textit{modality missing} and \textit{label scarcity}. For antibody-antigen pairings without structural information, many methods cannot be used to predict AAI. For instance, single residue mutations, which are crucial for AAI, often lack corresponding structural data. Directly replacing the mutated residues \cite{jiang2023dgcddg} will reduce the prediction accuracy. Moreover, due to the scarcity of labels, the research about supervised learning for AAI prediction  \cite{tsuruta2023avida,jin2024unsupervised} is minimal. Hence, there’s a shortage of datasets offering antibody-antigen structures and interaction labels (affinity and neutralization).

To address these challenges, we first propose a \textit{multi-modality representations} framework to capture the intricate hierarchical relationships within antibodies and antigens. The multi-modality representations incorporate both the 3D structural information (including residue, backbone atom, and side-chain atom) and 1D sequence information (order of amino acids) of the protein, which are physically meaningful and biologically meaningful. Based on these \textbf{Mu}lti-moda\textbf{L}ity representations, we propose an \textit{\textbf{A}ntibody-\textbf{A}ntigen \textbf{I}nteraction \textbf{P}rediction (MuLAAIP)} method. On one hand, MuLAAIP utilizes a graph attention module to delineate structural relationships within a protein, producing graph-level structural representation. On the other hand, it employs a normalized adaptive graph convolution module to depict inter-protein relationships, yielding protein-level sequence representation. These structural and sequence representations are subsequently fused and fed into a shared multi-layer perception for interaction prediction. 

We also developed a comprehensive multi-modality benchmark to propel advancements in the AAI field, including sequence, structure, and interaction labels for antibody-antigen pairings. This benchmark consists of wild-type affinity dataset, mutant-type affinity dataset with mutant antibody-antigen pairings, Alphaseq affinity dataset featuring mutant antibodies created via artificial point mutations, and SARS-CoV-2 neutralization dataset. The key contributions are outlined below.
\begin{enumerate}
\item We propose a multi-modality representation framework for antibody and antigen, designed to integrate 3D structural insights at the residue, backbone, and side-chain levels, complemented by 1D sequence data.
\item We introduce a novel prediction framework for AAI, termed MuLAAIP, which decodes the complex interplay between and within proteins, leveraging both graph-level structural representations and sequence-based relations. 
\item We create a comprehensive multi-modality AAI benchmark with structural and sequence information, and provide four distinct labels: three types of affinity labels and neutralization labels.
\item Comprehensive experiments on established benchmarks highlight MuLAAIP's superior performance in predicting antibody-antigen interactions and uncovering the underlying mechanisms.
\end{enumerate}

\section{Related work}
In recent years, significant progress has been made in protein representation learning \cite{jing2021learning, yuan2022structure, gao2023hierarchical, jiang2023dgcddg, song2022learning,li2024vidta}. Typically, protein structures are characterized using contact maps that define residue-level rigid structures, yet they may offer a coarse-grained view of protein dynamics, indicating a promising area for further research \cite{wang2023learning}. For instance, the HIGH-PPI model \cite{gao2023hierarchical} enhances protein-protein interaction prediction by utilizing a hierarchical graph constructed from residue contact maps and handcrafted features. Furthermore, existing research lacks exploration of fine-grained multi-modality representation learning for antibody antigen interactions.

\section{Method}
\subsection{Multi-modality representation framework}
\begin{figure*}[!t]
   \centering
   \includegraphics[width=1.0\textwidth]{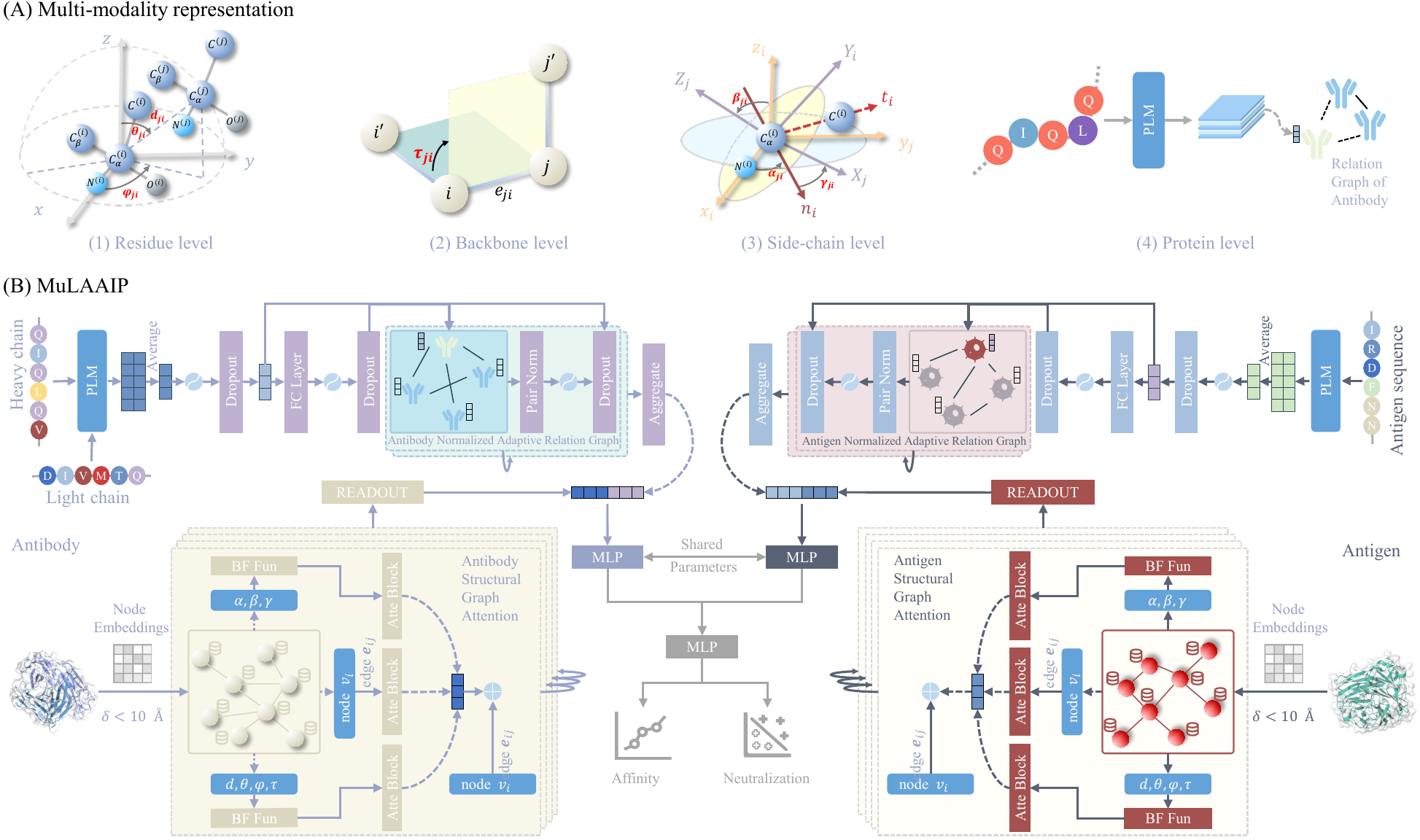}
   \caption{\textbf{Overview of multi-modality representation scheme and representation learning framework MuLAAIP.} (A) From coarse-to-fine, an antibody is represented using residue positions (\textit{i.e.}, $C_\alpha$ atom coordinates), backbone atom positions, side-chain atom positions, and sequence representation extracted from the PLM. (B) Depiction of the MuLAAIP framework.}
   \label{fig1}
   \vspace{-5mm}
\end{figure*}

We developed a comprehensive representation of antibodies and antigens, incorporating residue positions ($C_\alpha$ coordinates), backbone and side-chain atom positions, and sequence features derived from the protein language model (PLM) and sequence similarities from the relation graph, illustrated in Fig.~\ref{fig1} (A). We define a protein (antibody or antigen) with $n$ residue as $P=\{a^{(1)},a^{(2)},\ldots,a^{(n)}\}$. To establish a geometric representation of the protein structure, we initially depict the protein as a densely connected 3D graph $\mathcal{G}=(\mathcal{V},\mathcal{E},\mathcal{P})$. Specifically, $\mathcal{V}=\{\mathbf{v}_{i}\}_{i=1}^{n}$ denotes the node features, with individual residues acting as graph nodes. Each node comprises residue, backbone atom and side-chain atom embeddings. $\mathcal{E} = \{\mathbf{e}_{ij}\}_{i=1}^{n}{}_{j=1}^{n}$ signifies the edge features, where an edge exists between residues if the physical distance is within a predefined cutoff radius $c$. $\mathcal{P}=\{\mathbf{p}_{i}\}_{i=1}^{n}$ denotes the set of position matrices.

\subsubsection{3D Structural Representation}\label{subsubsec1}
\textbf{(1)Residue-level Representations.} For each residue node, we establish a local 3D spherical coordinate system (SCS) centered at its $C_{\alpha}$ atom coordinates,  using 3D coordinates $(d,\theta,\varphi)$, we unniquely define the position of each node with $d$ representing radial distance, $\theta$ denoting polar angle, and $\varphi$ indicating the azimuthal angle. The relative  position of node $j$ from the $i\operatorname{-th}$ node, with $d_{ij}$ as their  distance, $\theta_{ij}$ as the projection line angle on the $xy$ plane, and $\varphi_{ij}$ as the angle between the connecting line and the positive $z$-axis, is represented by the triplet $(d_{ij},\theta_{ij},\varphi_{ij})$. At the residue level, the geometric representation representation is captured in  $\mathcal{F}(G)_{\mathrm{a}} = \left\{(d_{ij}, \theta_{ij}, \phi_{ij}, \tau_{ij})\right\}_{i=1}^{n}{}_{j=1}^{\mathcal{N}_i}$, where $\tau_{ij}$ denotes the rotation angle for edge $\mathbf{e}_{ij}$, and $\mathcal{N}_i$ signifies the number of nodes connected to the $i\operatorname{-th}$ node, as depicted in Fig.~\ref{fig1} (A)-(1).

\textbf{(2) Backbone-level Representations.} In a standard residue, the backbone comprises three key atoms $(C,C_\alpha,N)$, which together define the residue structure. For the $i\operatorname{-th}$ node, these backbone atoms are denoted as $(C^{(i)},C_\alpha^{(i)},N^{(i)})$, with the local SCS centered at $C_\alpha^{(i)}$ as the origin, as depicted in Fig.~\ref{fig1} (A)-(2), formalized as :
\begin{small}
\begin{equation}
\begin{aligned}
\mathbf{x}_i &=\mathbf{r}_i^{N^{\left(i\right)}}-\mathbf{r}_i^{C_\alpha^{\left(i\right)}} \in\mathbb{R}^3 \\
\mathbf{t}_i&=\mathbf{r}_i^{C^{\left(i\right)}}-\mathbf{r}_i^{C_\alpha^{\left(i\right)}} \in\mathbb{R}^3 \\
\mathbf{z}_i&=\mathbf{x}_i\times\mathbf{t}_i \in\mathbb{R}^3 \\
\mathbf{y}_i &= \mathbf{x}_i \times \mathbf{z}_i \in\mathbb{R}^3 \\
\mathbf{n}_i&=\mathbf{z}_i \times \mathbf{z}_j \in\mathbb{R}^3
\end{aligned}
\end{equation}
\end{small}
Here, $\mathbf{r}_i^{C^{(i)}}, \mathbf{r}_i^{C_\alpha^{(i)}}, \mathbf{r}_i^{N^{(i)}}$ denotes the coordinates of atoms $C^{(i)},C_\alpha^{(i)},N^{(i)}$ respectively. $\mathbf{x}_i$ points from $\mathbf{r}_i^{C_\alpha^{(i)}}$ to $\mathbf{r}_i^{N^{(i)}}$, forming the basis vectors $\mathbf{x}_i, \mathbf{y}_i, \mathbf{z}_i$. When the backbone rotates by angle $\tau_{ij}$, we determine the coordinate axes $\mathbf{X}_j, \mathbf{Y}_j, \mathbf{Z}_j$ of the backbone. The intersection line $\mathbf{n}_i$ is found where the planes $\mathbf{X}_j{C_\alpha^{(i)}}\mathbf{Z}_j$ and $\mathbf{x}_i{C_\alpha^{(i)}}\mathbf{z}_i$ intersect. Hence, Euler angles $(\alpha_{ij},\beta_{ij},\gamma_{ij})$ are computed, with $\alpha_{ij}$ between $\mathbf{x}_i$ and the $\mathbf{n}_i$, $\beta_{ij}$ between $\mathbf{z}_i$ and $\mathbf{Z}_j$, and $\gamma_{ij}$ between $\mathbf{n}_i$ and $\mathbf{X}_j$. Overall, the backbone-level representation is denoted as $\mathcal{F}(G)_{\mathrm{b}} = \left\{(\alpha_{ij}, \beta_{ij}, \gamma_{ij})\right\}_{i=1, \, j=1}^{n, \, \mathcal{N}_i}$.

\textbf{(3) Side-chain-level Representations.} Assuming the constancy of bond lengths and angles within each residue node as rigid structures depicted in Fig.~\ref{fig1} (A)-(3), we underscore the significance of side-chain torsion angles in shaping structural characteristics \cite{wang2023learning}. We compute four torsion angles of the protein $\mathcal{F}(G)_{\mathrm{s}}=\left\{(\chi_i^1,\chi_i^2,\chi_i^3,\chi_i^4)\right\}_{i=1}^{n}$. These torsion angles are embedded on the 4-torus using the sine and cosine functions, specifically as $\begin{aligned}\{\sin,\cos\}\times(\chi_i^1,\chi_i^2,\chi_i^3,\chi_i^4)\end{aligned}$.

\subsubsection{1D Sequence Representation}\label{subsubsec2}
Upon noting a notable resemblance among antibody-antigen pairings with similar functional traits \cite{zhang2022predicting}, we devised an adaptive connectivity graph representation $\mathcal{F}(G)_{\mathrm{r}}=(\mathcal{V}_r,\mathcal{E}_r)$ depicted in Fig.~\ref{fig1} (A)-(4). Here $\mathcal{V}_r=\{\mathbf{h}_i\}_{i=1}^{n} $ represents the node embeddings, derived using a PLM that processes sequence data to yield sequence embeddings $\mathbf{h}_i$. The adjacency matrix $\mathcal{E}_r = \{\mathbf{e}_{ij}\}_{i=1, \, j=1}^{n, \, n}$ captures relationships between nodes. The entry $(i, j)$ in $\mathcal{E}_r$ denotes the cosine similarity between $\mathbf{h}_i$ and $\mathbf{h}_j$:
\begin{small}
    \begin{equation}
\mathbf{e}_{ij}=\frac{\mathbf{h}_i \cdot \mathbf{h}_j}{\|\mathbf{h}_i\|\|\mathbf{h}_j\|} 
\end{equation}
\end{small}

\subsection{MuLAAIP}
The proposed MuLAAIP architecture, illustrated in Fig.~\ref{fig1} (B), captures multi-modality representations of antibodies and antigens to predict AAI in an end-to-end manner. In the 3D structural graph attention module, ``BF Fun'' refers to feature encoding functions for geometric embedding (\textit{as detailed in the supplementary materials}), while ``Atte Block'' signifies the graph attention block. The 1D normalized adaptive  relation graph module is composed of relation graph, normalization, dropout, and activation function.

\subsubsection{3D Structural Graph Attention Module} 
Upon computing the multi-tier 3D graph representation, we acquire geometric structure embeddings encompassing residue-level, backbone-level, and side-chain-level representations. These geometric structure embeddings are individually input into graph attention blocks to obtain a structure-aware graph attention network:
\begin{small}
    \begin{equation}
\mathbf{v}_{i}^{(l+1)}=\sum_{j\in\mathcal{N}_i}\alpha_{ij}\mathbf{\Theta}_t\mathbf{v}_{j}^{(l)}
\end{equation}
\end{small}
where $\mathbf{\Theta}_{(\cdot)}$ denotes trainable parameter matrix, $\alpha_{ij}$ is the attention coefficients, which are computed as follows:
\begin{small}
    \begin{equation}
\alpha_{ij}=\frac{\exp(\sigma(\mathbf{a}_s\mathbf{\Theta}_s\mathbf{v}_i+\mathbf{a}_t\mathbf{\Theta}_t\mathbf{v}_j+\mathbf{a}_e\mathbf{\Theta}_e\mathbf{e}_{ij}))}{\sum_{k\in\mathcal{N}_i}\exp(\sigma(\mathbf{a}_s\mathbf{\Theta}_s\mathbf{v}_i+\mathbf{a}_t\mathbf{\Theta}_t\mathbf{v}_k+\mathbf{a}_e\mathbf{\Theta}_e\mathbf{e}_{ik}))}
\end{equation}
\end{small}
where $\sigma(\cdot)$ denotes the LeakyReLU nonlinear function, $\mathbf{a}_{(\cdot)}$ denotes the weight matrix of a fully connected (FC) layer. 

\noindent\textbf{Features Aggregation.} 
After aggregating node features from the graph attention blocks, we compute the graph-level structure representation $\mathbf{g}$ using the $\mathrm{READOUT}(\cdot)$ mechanism:
\begin{small}
    \begin{equation}
    \mathbf{g}=\mathrm{READOUT}(\{\mathbf{v}_{i}^{(l)}\}_{i=1}^{|n|})
\end{equation}
\end{small}
$\mathrm{READOUT}(\cdot)$ operation summarizes node features within a graph by aggregating them through summation along the node dimension.

\subsubsection{1D Normalized Adaptive Relation Graph Module} 
We utilize a PLM (\textit{i.e.,} TrotTrans) to extract sequence embeddings for each antibody. These embeddings undergo averaging, dropout, residual connections, and FC layers to derive the node embedding $\mathbf{h}$ for the antibody, along with the corresponding edges $\mathbf{e}_{ij}$. We then develop a normalized GCN to decipher protein-level representations. In this network, the $i\operatorname{-th}$ node's representations are updated from neighboring nodes $u\in \mathcal{N}(i)$ across $L\operatorname{-layer}$ layers of the GCN, considering adaptive connectivity relation representations $\mathcal{F}(G)_{\mathrm{r}}=(\mathcal{V}_r,\mathcal{E}_r)$:
\begin{small}
    \begin{equation}
    \mathbf{h}^{(l+1)}_i = \mathbf{\Theta}^{(l)} \sum_{j\in\{\mathcal{N}_i\cup i\}}
    \frac{\mathbf{e}^r_{ij}}{\sqrt{\hat{d}_j \hat{d}_i}} \mathbf{h}_j^{(l)}
\end{equation}
\end{small}
where $\hat{d}_i=1+ \sum_{j\in\mathcal{N}_i} \mathbf{e}^r_{ij}$, $\mathbf{\Theta}^{(l)}$ denotes the trainable parameter matrix. The heavy chain embedding and light chain embedding are concatenated together to get the protein representation $\mathbf{h}_G$. 

\noindent\textbf{Over-smoothing Alleviation.} 
Despite the GCN model embodying a specialized form of Laplacian smoothing among nodes' embeddings, it faces challenges of over-smoothing as the count of GCN layers escalates, this risk materializes particularly rapidly on small datasets with only a few convolutional layers. We adopt a two-step, center-and-scale, normalization procedure to alleviate this situation:
\begin{small}
    \begin{equation}
\tilde{H}_{c}^{(i)}=\tilde{H}^{(i)}-\frac{1}{N}\sum_{i=1}^{N}\tilde{H}^{(i)}
\end{equation}
\end{small}
\begin{small}
\begin{equation}
\hat{H}^{(i)}=s\cdot\frac{\tilde{H}_{c}^{(i)}}{\sqrt{\frac{1}{N}\sum_{i=1}^{N}\parallel\tilde{H}_{c}^{(i)}\parallel_{2}^{2}}}=s\sqrt{N}\cdot\frac{\tilde{H}_{c}^{(i)}}{\sqrt{\parallel\tilde{H}_{c}\parallel_{F}^{2}}}
\end{equation}
\end{small}
Here $\hat{H}$ is the normalized node representations, $\tilde{H}$ signifies the GCN output, and $s$ represents a constant hyperparameter governing the aggregate pairwise squared distance value.

\subsubsection{AAI Prediction Module} 
For an antibody-antigen pairing $(B^{(i)}, G^{(i)})$, we derive graph-level structural representations $\mathbf{g}_B^{(i)}$ and $\mathbf{g}_G^{(i)}$, along with sequence representations $\mathbf{h}_B^{(i)}$ and $\mathbf{h}_G^{(i)}$. The structural  embeddings and sequence embeddings of antibodies (and antigens) are jointly processed through a SMLP module. This architectural decision strategically reduces network parameters, enhancing computational efficiency while enabling robust feature integration. The consolidated representations are subsequently fed into the MLP to predict the interaction outcome.
\begin{small}
    \begin{equation}
p_i = \mathrm{MLP}(\mathbf{g}_B^{(i)} \| \mathbf{h}_B^{(i)} \|\mathbf{g}_G^{(i)} \|\mathbf{h}_G^{(i)}).
\end{equation}
\end{small}

The two subsequent tasks, binding affinity prediction and binary neutralization prediction, are handled separately. Their corresponding loss functions, $\mathcal{L}_{aff}$ and $\mathcal{L}_{neu}$, are formally delineated as:
\begin{small}
    \begin{equation}
    \begin{aligned}\mathcal{L}_{aff}&=\sum_{i\in\mathcal{V}}(y_{aff}^{(\mathrm{i})}-\hat{y}_{aff}^{(\mathrm{i})})^2\\&+\lambda_1\left\|\tilde{A_B}\right\|+\lambda_2\left\|\tilde{A_G}\right\|+\lambda\|\boldsymbol{W}\|^2\end{aligned}
\end{equation}
\end{small}

\begin{small}
    \begin{equation}
\begin{split}
\mathcal{L}_{neu}&=-\sum_{i\in\mathcal{V}}(y_{neu}^{(\mathrm{i})}\mathrm{ln}(\hat{y}_{neu}^{(\mathrm{i})})+(1-y_{neu}^{(\mathrm{i})})\mathrm{ln}(1-\hat{y}_{neu}^{(\mathrm{i})}))\\
&+\lambda_1\left\|\tilde{A_B}\right\|+\lambda_2\left\|\tilde{A_G}\right\|+\lambda\|\boldsymbol{W}\|^{2}    
\end{split}
\end{equation}
\end{small}
where $\tilde{A}$ is the adjancency matrix of the adaptive relation graph $\mathcal{F}(G)_{\mathrm{r}}$, $\left\|\tilde{A}\right\|$ denotes the sum of the absolute values in $\tilde{A}$. We take the $\left\|\tilde{A}\right\|$ and $\|\boldsymbol{W}\|^{2}$ as penalty terms to control the complexity of the model, $\lambda$, $\lambda_1$, and $\lambda_2$ are hyper-parameters for the trade-off for different loss components.

\section{Experiment results}
To validate our method, we tackle two pivotal inquiries: \textbf{Q1:} Does the MuLAAIP model excel in binding affinity and neutralization prediction tasks compared to state-of-the-art models? \textbf{Q2:} Does integrating sequence and structure data enhance performance, and how do diverse modalities influence interaction prediction efficacy?

\begin{table*}[htb]
\caption{Results on SARS-CoV-2 neutralization prediction task. The top two results are highlighted as $\textbf{1\textsuperscript{st}}$ and $\underline{2\textsuperscript{nd}}$. The performance metrics of PESI and PIPR are derived from \cite{wan2023pesi}.}
\vspace{-3mm}
\centering
\begin{tabular}{clccccc}
\hline
 &\textbf{Models}        & \textbf{ACC$\uparrow$}             & \textbf{F1$\uparrow$}              & \textbf{ROC-AUC$\uparrow$}         & \textbf{G-mean$\uparrow$}          & \textbf{MCC$\uparrow$}             \\ \hline
&DeepAAI       & $0.754_{\pm0.061}$ & $0.847_{\pm0.045}$ & $0.571_{\pm0.065}$ & $0.359_{\pm0.212}$ & $0.235_{\pm0.181}$ \\
  &PESI        & $0.741_{\pm0.017}$ & $0.840_{\pm0.011}$ & $0.631_{\pm0.069}$ & $0.215_{\pm0.120}$ & $0.124_{\pm0.094}$ \\
&AbAgIntPre    & $0.738_{\pm0.046}$ & $0.847_{\pm0.032}$ & $0.513_{\pm0.031}$ & $0.102_{\pm0.157}$ & $0.068_{\pm0.153}$ \\
\multirow{-4}{*}{\begin{tabular}[c]{@{}c@{}}Sequence-based\\ Deep Learning\end{tabular}}  
&PIPR          & $0.735_{\pm0.000}$ & $0.841_{\pm0.002}$ & $\underline{0.696}_{\pm0.058}$ & $0.014_{\pm0.019}$ & $0.001_{\pm0.002}$ \\ \hline
&ProtTrans      & $0.754_{\pm0.058}$ & ${0.849}_{\pm0.039}$ & $0.583_{\pm0.066}$ & $0.427_{\pm0.160}$ & $0.272_{\pm0.190}$ \\
&ESM2          & $0.748_{\pm0.065}$ & $0.843_{\pm0.047}$ & $0.582_{\pm0.041}$ & $\underline{0.582}_{\pm0.041}$ & $0.278_{\pm0.140}$ \\
&AbLang        & $0.729_{\pm0.052}$ & $0.832_{\pm0.043}$ & $0.547_{\pm0.045}$ & $0.344_{\pm0.183}$ & $0.135_{\pm0.145}$ \\
\multirow{-4}{*}{\begin{tabular}[c]{@{}c@{}}Protein Language\\ Based\end{tabular}}  
&BERT2DAb     & $0.745_{\pm0.054}$ & $0.848_{\pm0.037}$ & $0.539_{\pm0.043}$ & $0.236_{\pm0.202}$ & $0.137_{\pm0.154}$ \\ \hline
&ProtNet       & $0.667_{\pm0.141}$ & $0.717_{\pm0.227}$ & $0.617_{\pm0.083}$ & $0.509_{\pm0.176}$ & $\underline{0.284}_{\pm0.136}$ \\ 
&GVP-GNN       & $\underline{0.769}_{\pm0.041 }$ & $\underline{0.858}_{\pm0.026 }$ & $0.502 _{\pm0.014 }$ & $0.036 _{\pm0.040 }$ & $0.004 _{\pm0.002 }$ \\
\multirow{-3}{*}{\begin{tabular}[c]{@{}c@{}}Structure-based\\ Deep Learning\end{tabular}} 
&Atom3D-GNN       & $0.744 _{\pm0.030 }$ & $0.845 _{\pm0.018 }$ & $0.563 _{\pm0.041 }$  & $0.403 _{\pm0.101 }$ & $0.195 _{\pm0.074 }$ \\ \hline
  &GraphPPIS  & $0.734 _{\pm0.021 }$ & $0.842 _{\pm0.040 }$ & $0.512 _{\pm0.033 }$ & $0.206_{\pm0.110}$ & $0.003_{\pm0.005}$ \\
  &TAGPPI & $0.736_{\pm0.052}$ & ${0.847}_{\pm0.036}$& ${0.506}_{\pm0.080}$& ${0.000}_{\pm0.00}$& ${0.000}_{\pm0.00}$ \\
\multirow{-3}{*}{Co-modeling}  &MuLAAIP & $\textbf{0.816}_{\pm0.052}$ & $\textbf{0.874}_{\pm0.041}$ & $\textbf{0.757}_{\pm0.069}$ & $\textbf{0.728}_{\pm0.102}$ & $\textbf{0.549}_{\pm0.102}$ \\ \hline
\end{tabular}
\label{table1}
\vspace{-5mm}
\end{table*}

\subsection{Settings}
\textbf{Datasets.} To alleviate the scarcity of publicly available sequence-structure pairing, we have created a comprehensive benchmark and employed ESMFold \cite{lin2023evolutionary} to build the structures of uncharacterized proteins. The benchmark comprises four components, each featuring multi-modality antibody-antigen pairings with sequence-structure information: (1) The \textit{Wild-type} binding affinity dataset features 1,191 antibody-antigen pairings with binding affinity labels. (2) The \textit{Mutant-type} binding affinity dataset contains 1,742 antibody-antigen complexes featuring various mutations. (3) The \textit{Alphaseq} binding affinity dataset comprises 248,921 antibodies with binding affinities directed at a SARS-CoV-2 peptide. (4) The \textit{SARS-CoV-2} neutralization dataset comprises 310 antibody-antigen pairings labeled as 228 positive and 82 negative samples. \textbf{Benchmark development details can be found in the supplementary materials.}

\begin{figure}[htb]
\centering
\includegraphics[width=0.5\textwidth]{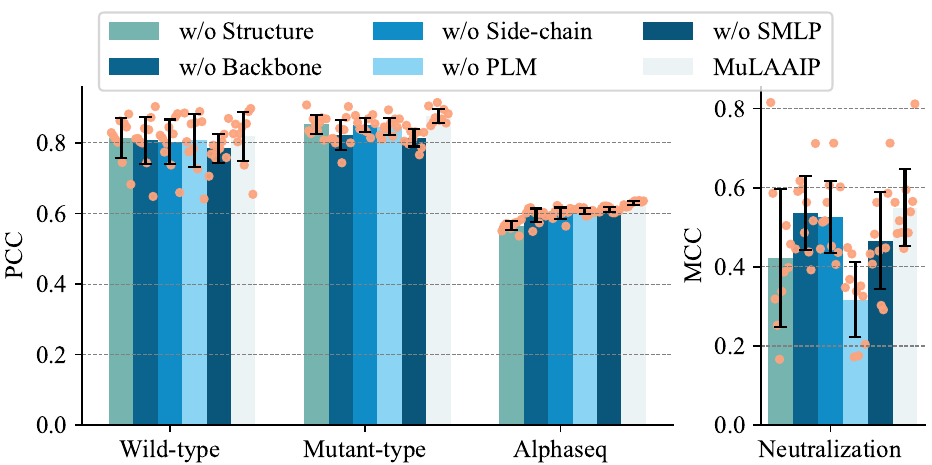}
\caption{Ablation results}
\label{fig2}
\vspace{-5mm}
\end{figure}

\textbf{Baselines and evaluation metrics.} We benchmark MuLAAIP against recent state-of-the-art baselines across various paradigms: (1) \textit{Sequence-based Deep Learning.} PIPR \cite{chen2019multifaceted} and DeepAAI \cite{zhang2022predicting} represent notable methods for affinity  and neutralization prediction. AREA-AFFINITY \cite{yang2023area} complements these in affinity prediction, while AbAgIntPre \cite{huang2022abagintpre} and PESI \cite{wan2023pesi} focus on AAI prediction from residue sequences. (2) \textit{Protein Language-based.} We select two antibody language models (ALMs), AbLang \cite{olsen2022ablang} and BERT2DAb \cite{luo2023bert2dab}, alongside two protein language models (PLMs), ProtTrans \cite{9477085} and ESM2 \cite{lin2023evolutionary}, as foundational benchmarks for antibody specificity modeling. (3) \textit{Structure-based Deep Learning.} ProtNet \cite{wang2023learning}, Atom3D \cite{townshend2021atomd}, and GVP-GNN \cite{jing2021learning} are key representations for protein structures. (4) \textit{Sequence-Structure co-modeling.} TAGPPI \cite{song2022learning} and GraphPPIS \cite{yuan2022structure} offer integrated approaches. Performance assessments involve MAE and PCC metrics for affinity prediction and adhere to the evaluation protocol in \cite{wan2023pesi} for neutralization prediction. The comparison results shown in Tables ~\ref{table1}-~\ref{table2}, the ablation study results shown in Fig.~\ref{fig2}.

\textbf{Experimental Setup.} We perform 10-fold validation for binding affinity and neutralization predictions, presenting average results and standard deviation. Experimental setups for protein language-based and structure-based deep learning methods align with prior works \cite{unsal2022learning, wang2023learning}. Other approaches align with prior work \cite{zhang2022predicting} using the Adam optimizer with a learning rate of $5e-5$, trained for 200 epochs with a batch size of 32 on a single GeForce RTX 4090 GPU. Training includes early stopping with a patience of 5.

\begin{table*}[htb]
\caption{Results of binding affinity prediction tasks. The top two results are highlighted as $\textbf{1\textsuperscript{st}}$ and $\underline{2\textsuperscript{nd}}$.}
\vspace{-3mm}
\centering
\begin{tabular}{lcccccc}
\hline
\multirow{2}{*}{\textbf{Models}} & \multicolumn{2}{c}{\textbf{Wild-type}} & \multicolumn{2}{c}{\textbf{Mutant-type}} & \multicolumn{2}{c}{\textbf{Alphaseq}} \\ \cline{2-7} 
                        & \textbf{MAE$\downarrow$}           & \textbf{PCC$\uparrow$}           & \textbf{MAE$\downarrow$}            & \textbf{PCC$\uparrow$}            & \textbf{MAE$\downarrow$}           & \textbf{PCC$\uparrow$}          \\ \hline
                        
DeepAAI        & $\underline{0.868}_{\pm0.152}$ & $\underline{0.792}_{\pm0.152}$  & $1.062_{\pm0.166}$ & $0.792_{\pm0.045}$ & $0.831_{\pm0.024}$ & $0.549_{\pm0.009}$ \\
AREA           & $1.587_{\pm0.196}$ & $0.348_{\pm0.089}$  & $1.752_{\pm0.560}$ & $0.590_{\pm0.063}$  & $1.061_{\pm0.035}$ & $0.315_{\pm0.014}$ \\
PIPR           & $1.012_{\pm0.112}$ & $0.694_{\pm0.083}$  & $1.162_{\pm0.103}$ & $0.723_{\pm0.043}$ & $1.001_{\pm0.005}$ & $0.400_{\pm0.002}$ \\ \hline
ProtTrans      & $0.909_{\pm0.134}$ & $0.759_{\pm0.064}$ & $0.912_{\pm0.077}$ & $0.827_{\pm0.022}$ & $0.772_{\pm0.006}$ & $\underline{0.618}_{\pm0.006}$ \\
ESM2          & $0.951_{\pm0.128}$ & $0.736_{\pm0.076}$ & $1.106_{\pm0.132}$ & $0.738_{\pm0.059}$ & $1.108_{\pm0.006}$ & $0.138_{\pm0.014}$ \\
AbLang         & $0.984_{\pm0.117}$ & $0.733_{\pm0.052}$ & $\underline{0.910}_{\pm0.075}$ & $\underline{0.830}_{\pm0.026}$ & $0.785_{\pm0.004}$ & $0.612_{\pm0.006}$ \\
BERT2DAb      & $0.976_{\pm0.105}$ & $0.700_{\pm0.008}$ & $1.062_{\pm0.061}$ & $0.779_{\pm0.032}$ & $0.845_{\pm0.003}$ & $0.549_{\pm0.005}$\\ \hline
ProtNet        & $0.989_{\pm0.117}$ & $0.737_{\pm0.072}$ & $1.029_{\pm0.147}$ & $0.800_{\pm0.040}$ & $0.790_{\pm0.030}$ & $0.596_{\pm0.002}$ \\ 
GVP-GNN        &  $1.243_{\pm0.123 }$ & $0.683 _{\pm0.050 }$  &  $1.471 _{\pm0.042 }$ & $0.623 _{\pm0.234 }$& $\underline{0.758}_{\pm0.025 }$ & $ 0.608_{\pm 0.014}$ \\
Atom3D-GNN        & $ 1.352_{\pm0.073 }$ & $ 0.522_{\pm0.068 }$ & $ 1.20_{\pm0.068 }$ & $ 0.725_{\pm 0.019}$& $ 0.816_{\pm0.025 }$ & $ 0.518_{\pm0.021 }$ \\ \hline
GraphPPIS   & $1.009_{\pm0.133}$ & $0.655_{\pm0.057}$& $1.149_{\pm0.092}$& $0.749_{\pm0.055}$& $1.055_{\pm0.006}$& $0.492_{\pm0.015}$ \\
TAGPPI  & $1.555_{\pm0.143 }$ & $0.247_{\pm0.055 }$& $1.865_{\pm0.0964 }$& $0.327_{\pm 0.019}$& $1.980_{\pm0.078 }$& $0.167_{\pm0.026 }$ \\
MuLAAIP   & $\textbf{0.740}_{\pm0.113}$ & $\textbf{0.819}_{\pm0.073}$ & $\textbf{0.815}_{\pm0.067}$ & $\textbf{0.876}_{\pm0.002}$& $\textbf{0.739}_{\pm0.011}$ & $\textbf{0.629}_{\pm0.006}$ \\ \hline
\end{tabular}
\label{table2}
\vspace{-5mm}
\end{table*}

\subsection{Performance comparison with other models}
\textbf{Q1} Noteworthy observations include: (1) \textit{Generalizability:} MuLAAIP demonstrates notable advantages across all four benchmarks, especially excelling in the alphaseq binding affinity benchmark. Despite the absence of standard structure, the incorporation of the sequence module aids in capturing essential information, reducing reliance on a singular structural modality. The performance of general protein-protein interaction prediction models such as GraphPPIS and TAGPPI falls short when compared to the specialized DeepAAI model designed specifically for AAI prediction. The ALMs and the general PLMs exhibit comparable performance in AAI prediction. (2) \textit{Strengths of multi-modality:} Leveraging protein sequences and structures for distinct tasks proves beneficial. In predicting wild-type antibody-antigen binding affinity, our tailored modules for sequence and structure adeptly capture crucial details compared to prior approaches. For mutate-type affinity prediction, our structure module effectively learns mutations from protein structures, overcoming limitations in sequence module relation learning due to high sequence similarities in mutated antibodies. (3) \textit{Applicability to imbalanced data:} In the SARS-CoV-2 neutralization prediction task, MuLAAIP significantly outperforms baselines, showcasing superior adaptability to imbalanced dataset.

\subsection{Ablation experiment}
In the ablation experiments, we investigated the influence of varying levels of representation and different modules on model performance. Fig.~\ref{fig2} visually summarizes the findings from our ablation studies on established benchmarks. 

\textbf{Q2} The results show that, \textbf{\uppercase\expandafter{\romannumeral1} multimodal information improves the performance of our model}. 
(1) MuLAAIP's integration of sequence and structural information surpasses single-module approaches, enhancing the fidelity of detailed representation learning. The model's performance is diminished upon the removal of either the structural graph module or the relational graph module. (2) MuLAAIP's multi-modality information assimilation, blending sequence and structural insights, reduces dependence on singular modalities and mitigates associated biases. Nevertheless, 'w/o PLM' exhibits strong performance in affinity prediction, emphasizing the pivotal role of protein structure in function determination. In neutralization prediction, a synergistic modeling of sequence and structure significantly boosts efficacy, contrasting with diminished performance when relying solely on either aspect. (3) Across diverse datasets, our model showcases decreased performance without the Shared Multilayer Perceptron (SMLP) module, highlighting its role in amalgamating antibody and antigen features. This integration proves critical, as the embeddings of antibodies and antigens encapsulate essential information influencing their binding affinities, particularly at the interaction interface.

\textbf{\uppercase\expandafter{\romannumeral2} Different levels of representation hold different biophysical meanings.} In all four datasets, the removal of various hierarchical representations, such as `w/o Backbone' and `w/o Side-chain', consistently led to a decline in model performance. (1) The mutation of the antibody side chain affects its binding to the epitope. The absence of comprehensive side chain information hinders the model's ability to discern pivotal mutational effects. (2) Blocking viral infection of host cells requires recognition of key sites on the virus. The detailed backbone and atomic level representation provide accurate and physically meaningful spatial information. (3) In scenarios where structural information falls short of precision, notably when mutant structures exhibit inaccuracies inherent in ESMFold predictions, incorporating sequence information remains pivotal in augmenting the model's perception capabilities.

\section{conclusion}
This paper presents MuLAAIP for AAI prediction, a multi-modality representation learning framework merging 3D structural and 1D sequence data. This integration adeptly captures intricate hierarchical antibody relationships. Leveraging a graph attention network for structural dynamics and a normalized adaptive graph convolution network for inter-protein sequences enhances the comprehension of AAI. Our thorough multi-modality benchmark establishes an evaluation standard for AAI prediction. Extensive experiments underscore MuLAAIP's substantial superiority over current state-of-the-art techniques.

\section*{Acknowledgment}
This work is supported by the National Natural Science Foundation of China (Grant No.62202186), the Hubei Provincial Natural Science Foundation Project (NO. 2023AFB342) and the National Natural Science Foundation of China (Grant No.62372196). The computation is completed in the HPC Platform of Huazhong University of Science and Technology. Shengqing Hu is the corresponding author.

\bibliographystyle{IEEEbib}
\bibliography{icme2025references}

\clearpage
\appendix

\textbf{Feature Encode.} While the 3D coordinates $(d,\theta,\varphi)$ uniquely specify the position of a node, this raw format lacks a meaningful representation and cannot be learned by a neural network. To this end, we leveraged spherical Fourier-Bessel bases with polynomial radial envelope function \cite{gasteiger_dimenet_2020} to transform the raw geometric features ($d,\theta,\varphi,\tau,\alpha,\beta,\gamma$) into physically-grounded representations, which consist of three components:
\begin{small}
    \begin{equation}
    \tilde{e}_{\mathrm{RBF},n}(d)=\sqrt{\frac2c}\frac{\sin(\frac{n\pi}cd)}d
\end{equation}
\end{small}
\begin{small}
    \begin{equation}
    \tilde{e}_{\mathrm{SBF},\ell n}(d,a)=\sqrt{\frac{2}{c^{3}j_{\ell+1}^{2}(z_{\ell n})}}j_{\ell}(\frac{z_{\ell n}}{c}d)Y_\ell^0(a)
\end{equation}
\end{small}
\begin{small}
    \begin{equation}
    \tilde{e}_{\mathrm{TBF},\ell n}(d,\theta,\varphi)=\sqrt{\frac{2}{c^{3}j_{\ell+1}^{2}(z_{\ell n})}}j_{\ell}(\frac{z_{\ell n}}{c}d)Y_\ell^m(\theta,\varphi)
\end{equation}
\end{small}
where $j_{\ell}(\cdot)$ is the $\ell \operatorname{-order}$ spherical Bessel function, $z_{ln}$ is the $n\operatorname{-th}$ root of the $\ell \operatorname{-order}$ Bessel function, $Y_\ell^m(\cdot)$ is the $\ell \operatorname{-order}$ spherical harmonic function of degree $m$, $\ell \in [0,\cdots,M-1]$, $m \in [-\ell, \cdots, \ell]$, $n \in [1,\cdots,N]$, $M$ and $N$ denote the highest orders for the spherical harmonics and spherical Bessel functions, respectively. Formally, $d$ is encoded with $\tilde{e}_{\mathrm{RBF},n}$, $(d,a)$ is encoded with $\tilde{e}_{\mathrm{SBF},\ell n}$ and $a \in \{\tau,\alpha,\beta,\gamma\}$, $(d,\theta,\varphi)$ is encoded with $\tilde{e}_{\mathrm{TBF},\ell n}$.

\textbf{Construction of the Proposed Benchmark.} Due to the lack of availability of high-quality structures and labels (affinity and neutralization) of AAI prediction, we have developed four datasets:
\begin{itemize}
\item \noindent\textbf{Wild-type Binding Affinity.} \textit{A comprehensive wild-type binding affinity benchmark, comprising a total of 1,191 curated antibody-antigen complex pairs.} The wild-type data utilized in our research was sourced from two widely-adopted dataset - namely, the SAbDab \cite{wilton2018sdab}  and the expanded antibody benchmark \cite{guest2021expanded}. The SAbDab dataset contained 1,274 antibodies annotated with Gibbs free energy ($\Delta G$) measurements. Complementing this, the expanded antibody benchmark comprised 51 unique antibody-antigen complex structures. Consistent with prior studies \cite{chen2019multifaceted,yang2023binding} on the task of antibody-antigen binding affinity prediction, we implemented a rigorous data de-duplication and cleansing protocol, eliminating overlaps and incomplete/erroneous entries, particularly those lacking essential antigen sequence data or exhibiting inconsistencies in atomic representations.

\item \noindent\textbf{Mutant-type Binding Affinity.} \textit{A comprehensive dataset of side-chain level mutations, comprising a total of 1,742 antigen-antibody complexes.} Analogous to our wild-type data curation, we pruned instances with missing information (\textit{e.g.}, absent antibody chains). Departing from standard sequence-to-affinity methods, we employed the ESMFold \cite{lin2023evolutionary} model for structure prediction. For antibody-antigen complexes with single or multiple mutations, we utilized ESMFold to forecast the mutated side-chain structures, preserving wild-type structures for unaffected regions. Mutant-type binding affinity are derived from the SKEMPI 2.0 \cite{jankauskaite2019skempi} and AB-Bind \cite{sirin2016ab} databases, with unified affinity labels represented by $\Delta G$.  
The AB-Bind \cite{sirin2016ab} dataset contains 709 mutations for which the binding affinity, as quantified by the change in binding affinity ($\Delta \Delta G$), has been experimentally determined. The $\Delta \Delta G$ value is computed as the difference between the mutant's ($\Delta G_{mut}$) and the wild-type's ($\Delta G_{wt}$) binding affinities:
\begin{equation}
\Delta \Delta G = \Delta G_{mut}-\Delta G_{wt} 
\end{equation}

The SKEMPI 2.0 \cite{jankauskaite2019skempi} dataset includes 1,211 entries, whose affinity are labeled in the form of dissociation constants ($K_D$), which adhere to the thermodynamic relationship \cite{jankauskaite2019skempi}:
\begin{equation}
    \Delta G= RT\ln{K_D}
\end{equation}
where $R$ denotes the gas constant and $T$ represents the reaction temperature in Kelvin. We convert all affinity labels into $\Delta G$.

\item \noindent\textbf{Alphaseq Binding Affinity.} \textit{A benchmark encompassing the binding affinity of 248,921 antibodies towards a SARS-CoV-2 peptide.} Alphaseq \cite{engelhart2022dataset} contains antibody mutants generated by artificial point mutations, and the mutants assigned to the test set are ``unseen''. We leveraged the ESMFold model to predict both antibodies and their target peptides, augmenting the benchmark with the intricate structural profiles of antibodies. The antigen featured in the Alphaseq dataset is a conserved peptide within the HR2 domain of the SARS-CoV-2 spike protein, specifically targeting the PDVDLGDISGINAS residue sequence. The antibodies in Alphaseq manifest as k-point mutated single-chain variable fragments (scFvs). 

\item \noindent\textbf{SARS-CoV-2 Neutralization.} \textit{A neutralization prediction benchmark for a total of 310 antigen-antibody pairs,228 pairs of positive samples and 82 pairs of negative samples.} Utilizing insights from previous research \cite{wan2023pesi,wu2024psc}, we incorporated both the structures and sequences of antibody-antigen pairs. This approach enables us to capture global residue interaction patterns, enhancing the understanding of the complex dynamics between the antibody and antigen. We use the protein structure with PDB code 7VXF to represent the structure of the antigen. 
\end{itemize}
In our wild-type dataset, the structures are exclusively derived from experimental determinations. For the mutant-type datasets, a subset of structures is predicted using the ESMFold model. In the Alphaseq dataset, all antibody structures are predicted by the ESMFold model

Due to the presence of non-canonical amino acids in certain antibodies or antigens, $C_\alpha$ atoms may be absent. Therefore, we replaced the spatial coordinates of the $C_\alpha$ atom with those of the N atom.

ESMFold \cite{lin2023evolutionary} runs on an A100-SXM-80G GPU with default parameters to predict antibody and antigen structures, maintaining a dataset-specific cutoff distance of 10 $\text{Å}$ as established in existing studies \cite{gao2023hierarchical, wang2023learning}.

\label{sec:supplement}

\end{document}